\documentclass[a4paper,twoside]{article}

\usepackage{epsfig}
\usepackage{subcaption}
\usepackage{calc}
\usepackage{amssymb}
\usepackage{amstext}
\usepackage{amsmath}
\usepackage{amsthm}
\usepackage{multicol}
\usepackage{pslatex}
\usepackage{apalike}
\usepackage{algorithm2e}
\usepackage{adjustbox}
\usepackage{booktabs}
\usepackage{arydshln}
\usepackage{url}
\usepackage{multirow}
\usepackage{todonotes}
\usepackage{color,xcolor,colortbl}
\usepackage{subcaption}
\usepackage{pgfplots}
\usepgfplotslibrary{fillbetween}
\usetikzlibrary{patterns}
\usepackage[bottom]{footmisc}
\usepackage{SCITEPRESS}     % Please add other packages that you may need BEFORE the SCITEPRESS.sty package.
\pgfplotsset{compat=1.18}

\makeatletter
\def\adl@drawiv#1#2#3{%
        \hskip.5\tabcolsep
        \xleaders#3{#2.5\@tempdimb #1{1}#2.5\@tempdimb}%
                #2\z@ plus1fil minus1fil\relax
        \hskip.5\tabcolsep}
\newcommand{\cdashlinelr}[1]{%
  \noalign{\vskip\aboverulesep
           \global\let\@dashdrawstore\adl@draw
           \global\let\adl@draw\adl@drawiv}
  \cdashline{#1}
  \noalign{\global\let\adl@draw\@dashdrawstore
           \vskip\belowrulesep}}
\makeatother

\newcommand{\quotes}[1]{``#1''}

\definecolor{lightblue}{HTML}{5DA5DA}
\definecolor{lightorange}{HTML}{FAA43A}
\definecolor{lightgreen}{HTML}{60BD68}
\definecolor{lightpurple}{HTML}{B276B2}

\begin{document}

\title{Evaluating Prompting Strategies for Zero-Shot Multilingual Aspect-Based Sentiment Analysis with Large Language Models}

\author{\authorname{Jakub \v{S}m\'{i}d\sup{1,2}\orcidAuthor{0000-0002-4492-5481} and Pavel Kr\'{a}l\sup{1,2}\orcidAuthor{0000-0002-3096-675X}}
\affiliation{\sup{1}Department of Computer Science and Engineering, University of West Bohemia in Pilsen, Univerzitn\'{i}, Pilsen, Czech Republic}
\affiliation{\sup{2}NTIS - New Technologies for the Information Society, University of West Bohemia in Pilsen, Univerzitn\'{i}, Pilsen, Czech Republic}
\email{\{jaksmid, pkral\}@fav.zcu.cz}
}


\abstract{While large language models (LLMs) have advanced natural language understanding, their ability to perform complex target-aspect-sentiment detection (TASD) -- a key task within aspect-based sentiment analysis (ABSA) -- in zero-shot multilingual settings remains underexplored. Existing ABSA research predominantly focuses on English or relies on fine-tuned task-specific models, leaving a gap in understanding how open-weight LLMs handle compound sentiment triplet extraction across diverse language families and domains. We present a systematic evaluation of zero-shot multilingual TASD across 12 languages and three domains using seven instruction-tuned open-weight models -- ranging from 8B to 35B parameters across dense and mixture-of-experts (MoE) architectures -- under four prompting paradigms: monolithic, chain-of-thought, sequential, and sequential refine. Our results show that direct monolithic prompting achieves strong zero-shot performance, outperforming multi-step pipelines, while dense models consistently outperform sparse MoE architectures. Through fine-grained error taxonomy profiling and step-level pipeline tracking, we demonstrate that initial aspect term recall forms a hard bottleneck, whereas multi-step decomposition degrades extraction quality; iterative refinement only partially recovers this loss on larger models while inducing net-negative over-refinement on smaller architectures. Finally, generation volume analysis shows that multi-step strategies incur up to $2.38\times$ token overhead without F1 gains, establishing monolithic prompting as the most effective and efficient paradigm for multilingual ABSA.}

\onecolumn \maketitle \normalsize \setcounter{footnote}{0} \vfill

\section{\uppercase{Introduction}}

Aspect-based sentiment analysis (ABSA) is a fine-grained natural language processing (NLP) task that aims to extract detailed sentiment information from text~\cite{absa}. ABSA includes several sentiment elements, e.g. aspect term ($a$), aspect category ($c$), and sentiment polarity ($p$). For example, in the review \textit{\quotes{I did enjoy the soup}}, the elements are \textit{\quotes{soup}}, \textit{\quotes{food quality}}, \textit{\quotes{positive}}, respectively. Implicit aspect terms, like in \textit{\quotes{Delicious!}} are often denoted as \textit{\quotes{null}}. Target-aspect-sentiment detection (TASD)~\cite{tasd} simultaneously extracts all three sentiment elements. In contrast, related subtasks like aspect term extraction target aspect terms alone, whereas E2E-ABSA~\cite{pontiki-etal-2014-semeval} extracts target terms alongside sentiment polarities.

While mainstream ABSA research primarily focuses on English, real-world applications require robust multilingual capabilities~\cite{wu-etal-2025-absa,pontiki-etal-2016-semeval}. However, annotated ABSA resources in non-English languages are scarce, and manual dataset creation is costly and labour-intensive. A common solution is cross-lingual ABSA, where a model is fine-tuned on a high-resource language (e.g. English) and evaluated on target languages. Nevertheless, syntactic shifts and semantic divergence across language families often degrade cross-lingual transfer performance. To mitigate these disparities, prior work relies on machine translation pipelines~\cite{zhang-etal-2021-cross} or cross-lingual data augmentation~\cite{smid-etal-2025-laca}. Furthermore, most multilingual ABSA literature focuses on simplified sub-tasks (e.g. ATE or E2E-ABSA), leaving full TASD tuple prediction under-explored in non-English contexts~\cite{SMID2025103073}.

An alternative paradigm that eliminates task-specific fine-tuning is the deployment of large language models (LLMs). Trained on massive multilingual corpora, instruction-tuned LLMs exhibit strong zero-shot capabilities across diverse NLP benchmarks. However, in complex ABSA tasks, zero-shot LLMs frequently underperform smaller, task-specific fine-tuned models -- largely due to the strict multi-variable constraints required by tuple extraction. Moreover, existing LLM evaluations in ABSA focus predominantly on English~\cite{smid-etal-2024-llama,zhang-etal-2024-sentiment}. Multilingual evaluations remain sparse and frequently rely on legacy LLM architectures~\cite{smid2025icaart,wu2024evaluatingzeroshotmultilingualaspectbased}, trailing behind the rapid evolution of modern open-weight models. Furthermore, the comparative efficacy of dense models against sparse mixture-of-experts (MoE) architectures -- which route tokens dynamically to a subset of specialized expert subnetworks rather than utilizing all parameters per token to reduce inference compute -- remains largely unexplored under zero-shot multilingual constraints. While advanced prompting paradigms like chain-of-thought (CoT) reasoning~\cite{3600270.3602070} improve complex multi-step reasoning, their efficacy in zero-shot cross-lingual TASD remains minimally studied and often yields inconsistent gains over direct prompting.

In this work, we address these research gaps by presenting a systematic empirical benchmark of modern open-weight LLMs on cross-lingual TASD across three domains and 12 languages, evaluating four distinct prompting paradigms. Our main contributions are: 1) We perform an extensive empirical investigation of zero-shot TASD across 12 languages and three evaluation domains (hotel, laptop, restaurant), evaluating compound sentiment extraction across diverse language families. 2) We systematically evaluate seven modern open-weight instruction-tuned models -- spanning dense and MoE architectures up to 35B parameters -- across four prompting paradigms (monolithic, chain-of-thought, sequential, and sequential refine). 3) We demonstrate that dense models systematically outperform MoE models of comparable or larger total scale, attributing this to higher active per-token parameter capacity required for simultaneous multilingual alignment and strict target extraction. 4) We quantitatively trace intermediate pipeline behaviour, showing that initial aspect extraction establishes an unrecoverable recall ceiling, while multi-step decomposition underperforms direct prompting and iterative refinement yields net-negative trade-offs on smaller models without matching monolithic extraction on larger architectures. 5) We conduct fine-grained error and generation overhead analyses identifying recall bottlenecks, schema ambiguity, and token inefficiencies in zero-shot TASD.

\section{\uppercase{Related Work}}

Recent studies systematically demonstrate that zero-shot LLMs generally lag behind smaller, specialized models fine-tuned specifically for ABSA tuple extraction tasks~\cite{bai-etal-2024-compound,smid-etal-2024-llama,zhang-etal-2024-sentiment,smid2025icaart,wu2024evaluatingzeroshotmultilingualaspectbased,filip2024finetuningmultilinguallanguagemodels}. Early explorations~\cite{zhang-etal-2024-sentiment,bai-etal-2024-compound} focused primarily on simplified ABSA tasks in English, noting that the performance gap between general-purpose LLMs and task-specific smaller Transformer-based models~\cite{vaswani2017attention}  -- e.g. fine-tuned mT5~\cite{xue-etal-2021-mt5} or XLM-R~\cite{conneau-etal-2020-unsupervised} -- is significantly wider in ABSA than in general classification benchmarks.

Extending these investigations to multilingual settings, recent evaluations~\cite{fehle-etal-2026-zero,smid-etal-2024-llama,smid2025icaart,wu2024evaluatingzeroshotmultilingualaspectbased} confirm that task-specific fine-tuning yields superior F1 scores, though parameter-efficient fine-tuned LLMs represent a strong middle ground. In zero-shot contexts, one work~\cite{wu2024evaluatingzeroshotmultilingualaspectbased} compares prompting variations across multiple languages, observing that complex reasoning prompts like chain-of-thought often fail to outperform direct single-pass prompts. Concurrently, modern instruction-tuned LLMs have proven effective as automated data annotators~\cite{hellwig-etal-2026-llm}, raising the question of whether state-of-the-art open-weight models can narrow the performance gap on complex cross-lingual TASD without task-specific gradient updates.

\begin{figure*}[t!]
    \centering
    % Left Column: Monolithic + CoT (Vertically Centered)
    \begin{minipage}[c]{0.48\textwidth}
        \centering
        \begin{subfigure}[b]{\linewidth}
            \centering
            \includegraphics[width=\linewidth]{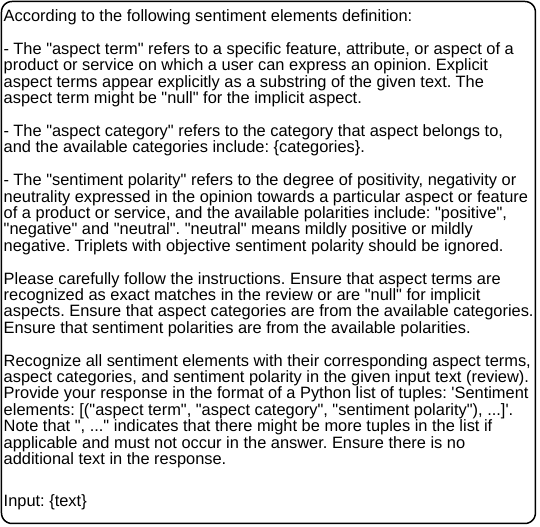}
            \caption{Monolithic Prompt}
            \label{fig:mono}
        \end{subfigure}
        
        \vspace{2.0ex} % Adjust vertical spacing between Mono and CoT if needed
        % \vspace{-0.5ex}
        \begin{subfigure}[b]{\linewidth}
            \centering
            \includegraphics[width=\linewidth]{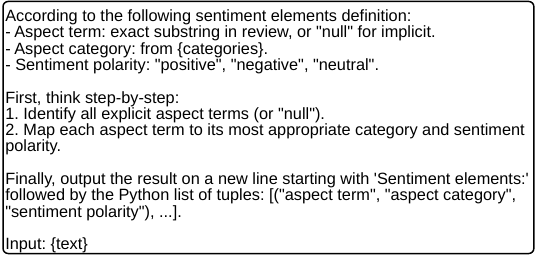}
            \caption{Chain-of-Thought (CoT) Prompt}
            \label{fig:cot}
        \end{subfigure}
    \end{minipage}%
    \hfill
    % Right Column: Multi-stage Sequential (Vertically Centered)
    \begin{minipage}[c]{0.48\textwidth}
        \centering
        \begin{subfigure}[b]{\linewidth}
            \centering
            \includegraphics[width=\linewidth]{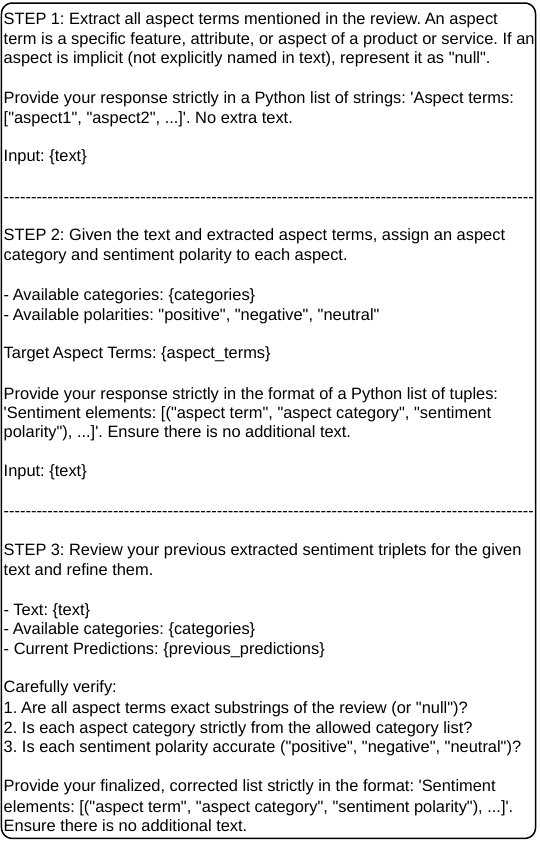}
            \caption{Multi-Stage Sequential Workflow}
            \label{fig:sequential}
        \end{subfigure}
    \end{minipage}

    % \caption{Prompting strategy architectures evaluated in this work: (a) single-pass direct monolithic prompt interface specifying sentiment definitions, negative guardrails, and strict output tuple constraints; (b) in-context chain-of-thought (CoT) prompt structure guiding step-by-step aspect-category-sentiment mapping; and (c) multi-stage sequential workflow separating candidate aspect extraction (Steps 1--2) from category mapping and iterative refinement (Step 3).}
    \caption{Prompting strategy architectures evaluated in this work: (a) single-pass direct monolithic prompt specifying sentiment definitions, negative guardrails, and output tuple constraints; (b) in-context chain-of-thought (CoT) prompt guiding step-by-step aspect-category-sentiment mapping; and (c) multi-stage sequential workflow separating aspect extraction (Step 1), category and polarity assignment (Step 2), and iterative tuple verification and refinement (Step 3).}
    \label{fig:prompts_overview}
\end{figure*}

\section{\uppercase{Experimental Setup}}

We evaluate seven instruction-tuned open-weight models spanning diverse parameter scales, release generations, and architectural paradigms -- including dense and sparse MoE designs (where only a subset of parameters is activated per token): LLaMA~3.1~8B (\texttt{LLaMA-3.1-8B-Instruct})~\cite{dubey2024llama3herdmodels}; Gemma~4 12B (\texttt{gemma-4-12B-it}, dense), Gemma~4 26B (\texttt{gemma-4-26B-A4B-it}, MoE with 4B active parameters), and Gemma~4 31B (\texttt{gemma-4-31B-it}, dense)~\cite{gemmateam2026gemma4technicalreport}; Qwen~3.6 27B (\texttt{Qwen3.6-27B}, dense)~\cite{qwen3.6-27b} and Qwen~3.6 35B (\texttt{Qwen3.6-35B-A3B}, MoE with 3B active parameters)~\cite{qwen36_35b_a3b}; and Qwen~3.8 27B (\texttt{Qwen3.8-27B}, dense)~\cite{qwen38}. All models are evaluated in their default inference configurations: Qwen-3.6 and Qwen-3.8 operate with native internal thinking mode enabled by default, whereas Gemma~4 models execute with thinking mode disabled by default, and LLaMA-3.1-8B lacks native reasoning traces. Focusing on locally deployable open-weight models avoids proprietary API costs while ensuring full reproducibility.

All models are implemented using the HuggingFace Transformers library~\cite{wolf-etal-2020-transformers} and deployed via the vLLM inference engine~\cite{kwon2023efficient} for high-throughput batch generation. To balance memory efficiency with empirical fidelity, models up to 26B parameters are evaluated in native half-precision (BF16/FP16), whereas larger models ($\ge 27$B parameters) utilize 4-bit floating-point (NVFP4) quantization. All inferences run with greedy decoding on a single NVIDIA H100 NVL GPU (96 GB VRAM).

We evaluate the TASD task across three core domains (restaurant, hotel, and laptop) using a 11-language subset of the parallel M-ABSA benchmark~\cite{wu-etal-2025-absa}: English (en), Spanish (es), French (fr), Dutch (nl), Russian (ru), Turkish (tr), Japanese (ja), Korean (ko), Vietnamese (vi), Thai (th), and Chinese (zh). Unlike SemEval-2016~\cite{pontiki-etal-2016-semeval}, which covers fewer languages primarily in the restaurant domain with non-parallel splits, this parallel benchmark translates English reviews into target languages, holding semantic content and label distributions strictly constant across language families. Because M-ABSA relies on translated non-English text and was released in early 2025 (introducing potential translationese artifacts and pre-training data contamination), we supplement it with the natively annotated Czech (cs) Restaurant dataset~\cite{smid-etal-2024-czech,md-etal-2026-extending}. Table~\ref{tab:data} details the dataset statistics for the evaluated test splits.

\begin{table}[ht!]
    \centering
    \caption{Data statistics for datasets used.}
    \begin{adjustbox}{width=0.85\linewidth}   
        \begin{tabular}{@{}lrrrr@{}}
        \toprule
          & \textbf{Restaurant (cs)}                    &  \textbf{Restaurant}   & \textbf{Hotel} & \textbf{Laptop} \\ \midrule
          No. sentences & 616 & 522 & 584         & 532            \\
          No. aspects & 1,390 & 975 & 714         & 610           \\ \bottomrule 
        \end{tabular}
    \end{adjustbox}
    \label{tab:data}
\end{table}

We use micro F1-score as the primary evaluation metric following prior work, counting a predicted tuple as correct only if its aspect term, category, and polarity exactly match the gold annotation.

\begin{table*}[ht!]
\centering
\caption{Micro-F1 TASD scores compared to the zero-shot fine-tuned mT5 baseline~\cite{wu-etal-2025-absa} (monolingual for English, respectively). For each model, language, and domain (Dom.), the best result is \underline{underlined}. For each language and domain, the best overall results is in \textbf{bold}. MO = monolithic prompt, COT = chain-of-thought prompt, SQ = sequential pipeline prompt, SR = sequential refined prompt. Models denoted with * are MoE architectures.}
\begin{adjustbox}{width=\linewidth}
\begin{tabular}{@{}llcccccccccccccccccccccccc@{}}
\toprule
\multicolumn{26}{c}{\textit{Indo-European Languages}}                                                                                                                                                                                                                                                                                                                                                                                                         \\ \midrule
\multirow{2}{*}{\rotatebox[origin=c]{90}{\begin{tabular}[c]{@{}c@{}}\textbf{Dom.}\end{tabular}}}  &                                   & \multicolumn{4}{c}{cs}                                                  & \multicolumn{4}{c}{en}                                                        & \multicolumn{4}{c}{es}                                & \multicolumn{4}{c}{fr}                                & \multicolumn{4}{c}{nl}                                & \multicolumn{4}{c}{ru}                                   \\ \cmidrule(lr){3-6} \cmidrule(lr){7-10} \cmidrule(lr){11-14} \cmidrule(lr){15-18} \cmidrule(lr){19-22} \cmidrule(lr){23-26}
             & \textbf{Model}                    & MO             & COT            & SQ             & SR                   & MO                   & COT            & SQ                   & SR             & MO             & COT            & SQ          & SR    & MO             & COT            & SQ          & SR    & MO             & COT            & SQ          & SR    & MO             & COT         & SQ          & SR          \\ \midrule
\multirow{7}{*}{\rotatebox[origin=c]{90}{\begin{tabular}[c]{@{}c@{}}hotel\end{tabular}}}      & Gemma 4 12B             &   --             &      --          &    --            &     --                 & \underline{40.34}          & 33.13          & 34.91                & 32.10          & \underline{38.68}    & 26.00          & 29.43       & 27.36 & \underline{36.51}    & 23.90          & 25.95       & 23.48 & \underline{40.52}    & 28.02          & 31.13       & 29.31 & \underline{40.74}    & 32.77       & 30.09       & 29.54       \\
                            & Gemma 4 26B*         &     --           &       --         &        --        &      --                & \underline{38.80}          & 37.41          & 31.71                & 33.58          & \underline{34.32}    & 28.18          & 26.15       & 28.04 & \underline{30.27}    & 27.61          & 23.39       & 24.85 & \underline{38.08}    & 31.87          & 28.50       & 30.90 & 39.45          & \underline{39.54} & 31.19       & 32.04       \\
                            & Gemma 4 31B       &     --           &        --        &      --          &    --                  & \underline{47.10}          & 41.39          & 40.32                & 41.08          & \underline{38.94}    & 34.34          & 36.51       & 37.49 & 32.57          & 30.08          & \underline{33.14} & 32.61 & \underline{40.82}    & 36.00          & 37.74       & 37.55 & \underline{44.89}    & 43.82       & 40.17       & 40.72       \\
                            & LLaMA 3.1 8B      &        --        &      --          & --      &          --            & 5.37                 & 10.19 & \underline{21.86} & 19.40 & 6.63           & 14.81          & \underline{21.45} & 19.10 & 4.52           & 12.00          & \underline{20.66} & 18.78 & 6.23           & 11.75          & \underline{21.70} & 18.60 & 5.16           & 11.52       & \underline{21.33} & 18.80       \\
                            & Qwen 3.6 27B          &    --            &     --           &      --          &        --              & \textbf{49.32}       & 46.36          & 42.85                & 46.41          & \textbf{44.54} & 44.48          & 38.54       & 43.16 & \textbf{40.60} & 38.74          & 35.59       & 37.84 & \textbf{46.53} & 44.07          & 39.70       & 45.14 & \textbf{47.16} & 46.19       & 38.20       & 40.97       \\
                            & Qwen 3.6 35B* &        --        &       --         &  --              &   --                   & \underline{43.33}          & 43.12          & 40.65                & 42.00          & 38.27          & \underline{38.77}    & 36.73       & 38.22 & 35.25          & 33.71          & \underline{35.51} & 34.97 & \underline{40.37}    & 39.13          & 37.75       & 38.16 & \underline{42.31}    & 41.87       & 36.88       & 39.06       \\
                            & Qwen 3.8 27B          &   --             &    --            &     --           &   --                   & \underline{44.84} & 44.53          & 36.85                & 40.58          & \underline{40.30}    & 40.00          & 32.28       & 36.18 & \underline{38.31}    & 36.41          & 30.16       & 32.97 & \underline{43.09}    & 40.63          & 32.94       & 37.30 & \underline{44.08}    & 42.98       & 33.01       & 38.39       \\ \cdashlinelr{2-26}
& \cite{wu-etal-2025-absa} &   \multicolumn{4}{c}{--} & \multicolumn{4}{c}{40.69} & \multicolumn{4}{c}{36.77} & \multicolumn{4}{c}{19.49} & \multicolumn{4}{c}{26.30} & \multicolumn{4}{c}{20.57} \\                          
                            \cdashlinelr{1-26}
\multirow{7}{*}{\rotatebox[origin=c]{90}{\begin{tabular}[c]{@{}c@{}}laptop\end{tabular}}}     & Gemma 4 12B             &   --             &    --            &      --          &      --                & \underline{27.84}          & 25.54          & 21.16                & 22.78          & \underline{26.06}    & 25.00          & 18.83       & 19.81 & \underline{24.55}    & 21.97          & 17.69       & 18.56 & \underline{28.75}    & 24.14          & 20.38       & 21.16 & \underline{28.14}    & 24.87       & 19.23       & 20.91       \\
                            & Gemma 4 26B*         & --      & --      &     --           & --            & \underline{27.61}          & 26.37          & 23.54                & 24.54          & \underline{24.93}    & 24.10          & 21.77       & 23.13 & \underline{26.17}    & 21.28          & 20.12       & 23.61 & \underline{29.45}    & 21.88          & 22.14       & 25.56 & \underline{28.13}    & 27.43       & 23.49       & 25.31       \\
                            & Gemma 4 31B       &    --            &     --           &   --             &  --                    & \underline{32.36}          & 30.31          & 28.82                & 27.91          & \underline{32.17}    & 28.69          & 26.32       & 27.52 & 26.82          & \underline{27.59}    & 25.71       & 26.36 & \underline{34.50}    & 30.84          & 29.91       & 30.13 & 32.36          & \underline{32.88} & 28.72       & 29.36       \\
                            & LLaMA 3.1 8B      &    --            &      --          &   --             &   --                   & 11.06                & 11.15          & \underline{14.94}          & 13.89          & 11.49          & 11.05          & \underline{13.81} & 11.19 & 9.22           & 9.08           & \underline{14.18} & 9.52  & 9.91           & 12.15          & \underline{16.30} & 12.80 & 10.53          & 10.91       & \underline{15.50} & 11.77       \\
                            & Qwen 3.6 27B          &       --         &     --           &     --           &    --                  & \textbf{36.94}       & 35.62          & 31.39                & 32.82          & \textbf{32.40} & 31.43          & 26.15       & 30.48 & 29.79          & \textbf{31.74} & 26.52       & 29.15 & 36.10          & \textbf{36.94} & 29.96       & 33.78 & \underline{34.25}    & 34.24       & 27.32       & 32.79       \\
                            & Qwen 3.6 35B* &        --        &      --          &   --             &    --                  & 33.33                & \underline{33.54}    & 30.40                & 30.54          & 30.89          & \underline{30.95}    & 27.72       & 29.01 & \underline{30.59}    & 29.87          & 26.49       & 28.43 & 33.33          & \underline{34.43}    & 30.14       & 30.61 & \textbf{35.68} & 34.52       & 27.86       & 29.73       \\
                            & Qwen 3.8 27B          &    --            & --      & --      & --            & \underline{34.91}          & 34.56          & 26.04                & 31.21          & \underline{29.98}    & 29.97          & 21.32       & 24.39 & \underline{29.01}    & 28.99          & 21.05       & 25.07 & 33.21          & \underline{33.40}    & 24.70       & 29.76 & 31.42          & \underline{34.55} & 22.86       & 27.52       \\  
                             \cdashlinelr{2-26}
& \cite{wu-etal-2025-absa} &   \multicolumn{4}{c}{--} & \multicolumn{4}{c}{43.54} & \multicolumn{4}{c}{22.46} & \multicolumn{4}{c}{24.91} & \multicolumn{4}{c}{33.54} & \multicolumn{4}{c}{18.37} \\      
                            \cdashlinelr{1-26}
\multirow{7}{*}{\rotatebox[origin=c]{90}{\begin{tabular}[c]{@{}c@{}}restaurant\end{tabular}}} & Gemma 4 12B             & \underline{58.15}    & 45.10          & 43.73          & 44.87                & \underline{57.66}          & 50.34          & 49.10                & 48.61          & \underline{53.39}    & 39.62          & 42.87       & 43.37 & \underline{48.59}    & 31.75          & 38.03       & 36.72 & \underline{55.22}    & 37.98          & 43.72       & 42.16 & \underline{57.75}    & 53.21       & 41.72       & 43.27       \\
                            & Gemma 4 26B*         & \underline{57.29}    & 51.15          & 49.96          & 51.09                & \underline{59.25}          & 53.68          & 49.97                & 50.00          & \underline{53.88}    & 38.99          & 44.55       & 44.65 & \underline{47.93}    & 38.63          & 38.91       & 37.92 & \underline{51.93}    & 41.54          & 43.19       & 43.08 & \underline{57.12}    & 54.97       & 47.04       & 48.39       \\
                            & Gemma 4 31B       & \textbf{63.25} & 58.22          & 58.38          & 59.46                & \underline{62.92}          & 59.37          & 57.00                & 56.18          & \underline{56.06}    & 50.06          & 52.65       & 51.97 & \underline{49.47}    & 43.28          & 45.77       & 45.64 & \underline{56.97}    & 50.29          & 54.58       & 53.23 & \underline{62.39}    & 58.63       & 55.62       & 57.25       \\
                            & LLaMA 3.1 8B      & 14.04          & 18.51          & \underline{33.72}    & 24.47                & 10.34                & 21.57          & \underline{36.92}          & 30.10          & 11.79          & 19.07          & \underline{30.10} & 25.78 & 7.39           & 17.76          & \underline{29.62} & 24.14 & 7.86           & 20.22          & \underline{34.80} & 26.37 & 7.76           & 21.01       & \underline{30.64} & 23.83       \\
                            & Qwen 3.6 27B          & 59.82          & 61.61          & 56.61          & \underline{62.11}          & \textbf{68.56}       & 66.02          & 60.26                & 64.06          & 60.07          & \textbf{61.08} & 54.94       & 57.20 & \textbf{56.52} & 55.00          & 50.32       & 51.25 & \textbf{62.54} & 61.10          & 55.74       & 59.34 & \textbf{62.86} & 62.50       & 54.83       & 57.70       \\
                            & Qwen 3.6 35B* & 56.72 & 57.22 & 53.65 & \underline{57.38} & \underline{62.90}          & 62.63          & 58.04                & 59.46          & \underline{57.56}    & 56.59          & 53.25       & 54.51 & \underline{51.81}    & 49.60          & 48.42       & 48.82 & \underline{60.88}    & 59.56          & 55.37       & 55.62 & \underline{61.73}    & 59.68       & 53.09       & 55.23       \\
                            & Qwen 3.8 27B          & \underline{57.64}    & 53.13          & 43.76          & 52.09                & \underline{63.78}          & 61.09          & 55.11                & 58.84          & \underline{59.04}    & 55.75          & 49.13       & 52.59 & \underline{53.43}    & 50.83          & 46.12       & 47.52 & \underline{58.83}    & 57.09          & 48.74       & 54.20 & \underline{59.29}    & 57.05       & 47.45       & 52.09       \\
                            \cdashlinelr{2-26}
& \cite{wu-etal-2025-absa} &   \multicolumn{4}{c}{--} & \multicolumn{4}{c}{48.87} & \multicolumn{4}{c}{23.83} & \multicolumn{4}{c}{26.83} & \multicolumn{4}{c}{25.27} & \multicolumn{4}{c}{23.36} \\ 
                            \midrule 
\multicolumn{26}{c}{\textit{Non-Indo-European Languages}}                                                                                                                                                                                                                                                                                                                                                                                                     \\ \midrule
\multirow{2}{*}{\rotatebox[origin=c]{90}{\begin{tabular}[c]{@{}c@{}}\textbf{Dom.}\end{tabular}}}                            &                                    & \multicolumn{4}{c}{tr}                                                  & \multicolumn{4}{c}{ja}                                                        & \multicolumn{4}{c}{ko}                                & \multicolumn{4}{c}{vi}                                & \multicolumn{4}{c}{th}                                & \multicolumn{4}{c}{zh}                                   \\ \cmidrule(lr){3-6} \cmidrule(lr){7-10} \cmidrule(lr){11-14} \cmidrule(lr){15-18} \cmidrule(lr){19-22} \cmidrule(lr){23-26}
            & \textbf{Model}                     & MO             & COT            & SQ             & SR                   & MO                   & COT            & SQ                   & SR             & MO             & COT            & SQ          & SR    & MO             & COT            & SQ          & SR    & MO             & COT            & SQ          & SR    & MO             & COT         & SQ          & SR          \\ \midrule
\multirow{7}{*}{\rotatebox[origin=c]{90}{\begin{tabular}[c]{@{}c@{}}hotel\end{tabular}}}      & Gemma 4 12B             & \underline{41.29}    & 33.27          & 32.33          & 31.36                & \underline{42.36}          & 34.31          & 32.59                & 31.20          & \underline{39.52}    & 30.96          & 32.64       & 30.34 & \underline{38.31}    & 31.59          & 31.20       & 30.06 & \underline{38.65}    & 34.15          & 29.40       & 28.90 & \underline{38.75}    & 32.98       & 32.70       & 30.45       \\
                            & Gemma 4 26B*         & \underline{39.14}    & 38.08          & 30.32          & 32.02                & 39.67                & \underline{39.96}    & 30.09                & 32.61          & 33.96          & \underline{35.89}    & 28.50       & 30.57 & 36.98          & \underline{37.18}    & 29.24       & 30.54 & 35.73          & \underline{38.91}    & 28.78       & 29.58 & 37.47          & \underline{38.90} & 30.24       & 31.25       \\
                            & Gemma 4 31B       & \textbf{47.44} & 44.56          & 39.91          & 42.34                & \underline{44.41}          & 41.21          & 40.21                & 41.09          & \underline{43.41}    & 40.47          & 37.34       & 38.17 & \underline{42.25}    & 40.38          & 39.64       & 40.43 & \textbf{43.89} & 40.29          & 38.16       & 38.92 & \underline{44.28}    & 41.05       & 39.74       & 39.62       \\
                            & LLaMA 3.1 8B      & 10.17          & 11.33          & \underline{20.17}    & 17.53                & 6.61                 & 12.02          & \underline{23.09}          & 19.81          & 7.22           & 9.92           & \underline{20.39} & 18.96 & 5.31           & 7.30           & \underline{21.95} & 20.46 & 8.33           & 10.22          & \underline{22.98} & 20.91 & 7.32           & 13.43       & \underline{24.54} & 21.64       \\
                            & Qwen 3.6 27B          & 46.36          & \underline{46.69}    & 39.72          & 46.12                & 47.19                & \textbf{47.87} & 40.50                & 46.14          & \textbf{43.55} & 43.09          & 38.33       & 42.11 & 43.13          & \textbf{43.82} & 40.53       & 42.39 & \underline{38.19}    & 37.56          & 32.52       & 38.05 & \textbf{47.31} & 45.43       & 41.05       & 44.80       \\
                            & Qwen 3.6 35B* & 42.30          & \underline{42.50}    & 37.86          & 39.32                & 40.88                & \underline{42.50}    & 37.99                & 39.17          & \underline{39.43}    & 38.38          & 37.34       & 39.22 & \underline{38.64}    & 38.49          & 37.40       & 38.36 & 28.00          & \underline{29.08}    & 26.56       & 26.82 & 40.40          & 39.42       & 38.98       & \underline{40.94} \\
                            & Qwen 3.8 27B          & \underline{42.83}    & 41.73          & 33.51          & 37.71                & \underline{44.09}          & 40.40          & 33.18                & 38.34          & \underline{40.09}    & 37.82          & 31.34       & 35.72 & \underline{39.96}    & 39.00          & 32.54       & 36.53 & \underline{32.84}    & 29.96          & 24.34       & 28.84 & \underline{42.29}    & 41.27       & 33.79       & 37.13       \\ 
                             \cdashlinelr{2-26}
& \cite{wu-etal-2025-absa} &   \multicolumn{4}{c}{17.71} & \multicolumn{4}{c}{25.30} & \multicolumn{4}{c}{13.36} & \multicolumn{4}{c}{15.69} & \multicolumn{4}{c}{20.90} & \multicolumn{4}{c}{24.45} \\
                            \cdashlinelr{1-26}
\multirow{7}{*}{\rotatebox[origin=c]{90}{\begin{tabular}[c]{@{}c@{}}laptop\end{tabular}}}     & Gemma 4 12B             & \underline{28.53}    & 26.62          & 17.43          & 19.02                & \underline{30.73}          & 26.91          & 20.28                & 21.01          & \underline{29.21}    & 25.42          & 17.29       & 18.83 & 24.18          & \underline{24.63}    & 17.29       & 18.36 & \underline{27.08}    & 25.45          & 17.03       & 18.14 & \underline{29.85}    & 26.49       & 18.15       & 19.51       \\
                            & Gemma 4 26B*         & \underline{26.52}    & 24.07          & 20.23          & 23.27                & \underline{28.45}          & 26.83          & 24.82                & 27.76          & 24.75          & \underline{26.26}    & 19.77       & 23.51 & 22.67          & \underline{23.78}    & 20.13       & 20.94 & 27.13          & \underline{28.23}    & 22.40       & 25.02 & \underline{27.83}    & 25.49       & 22.14       & 24.67       \\
                            & Gemma 4 31B       & \underline{33.29}    & 31.56          & 24.60          & 30.02                & \underline{34.44}          & 32.72          & 28.46                & 29.89          & \underline{29.11}    & 28.17          & 26.12       & 26.38 & \underline{28.65}    & 27.64          & 22.35       & 22.18 & 31.90          & \textbf{33.06} & 25.10       & 26.50 & \underline{31.14}    & 30.37       & 26.33       & 27.33       \\
                            & LLaMA 3.1 8B      & 11.77          & 8.29           & \underline{14.01}    & 9.97                 & 10.06                & 10.75          & \underline{17.79}          & 14.64          & 12.99          & 12.21          & \underline{16.15} & 13.42 & 8.61           & 7.52           & \underline{13.44} & 9.51  & 9.39           & 8.52           & \underline{16.23} & 12.83 & 12.98          & 13.49       & \underline{14.39} & 13.87       \\
                            & Qwen 3.6 27B          & 35.52          & \textbf{35.67} & 26.28          & 34.96                & 33.90                & \textbf{37.24} & 29.72                & 34.16          & 31.25          & \textbf{33.48} & 26.47       & 32.18 & 28.99          & \textbf{29.37} & 25.68       & 27.57 & 29.86          & \underline{31.17}    & 23.44       & 29.02 & \textbf{35.38} & 35.16       & 28.18       & 31.90       \\
                            & Qwen 3.6 35B* & 33.00          & \underline{35.08}    & 26.93          & 29.40                & 33.66                & \underline{34.05}    & 27.95                & 27.68          & 31.85          & \underline{32.70}    & 27.75       & 28.01 & \underline{27.95}    & 27.62          & 26.46       & 26.56 & 25.20          & \underline{27.16}    & 21.46       & 23.15 & 31.81          & \underline{32.66} & 28.06       & 29.64       \\
                            & Qwen 3.8 27B          & 30.45          & \underline{33.26}    & 22.39          & 26.96                & 29.47                & \underline{31.35}    & 20.72                & 26.23          & \underline{30.54}    & 29.15          & 21.18       & 25.54 & \underline{28.63}    & 28.59          & 21.27       & 23.37 & 20.43          & \underline{25.46}    & 14.75       & 20.57 & 30.60          & \underline{32.10} & 20.68       & 25.99       \\ 
                                                         \cdashlinelr{2-26}
& \cite{wu-etal-2025-absa} &   \multicolumn{4}{c}{20.88} & \multicolumn{4}{c}{22.22} & \multicolumn{4}{c}{18.01} & \multicolumn{4}{c}{19.59} & \multicolumn{4}{c}{22.56} & \multicolumn{4}{c}{22.55} \\
                            \cdashlinelr{1-26}
\multirow{7}{*}{\rotatebox[origin=c]{90}{\begin{tabular}[c]{@{}c@{}}restaurant\end{tabular}}} & Gemma 4 12B             & \underline{53.15}    & 45.89          & 42.45          & 43.93                & \underline{58.56}          & 51.44          & 47.17                & 47.07          & \underline{54.58}    & 49.78          & 42.36       & 42.50 & \underline{50.62}    & 47.55          & 40.78       & 39.61 & \underline{52.88}    & 51.26          & 39.29       & 40.63 & \underline{55.10}    & 49.16       & 45.00       & 43.94       \\
                            & Gemma 4 26B*         & \underline{52.99}    & 49.75          & 44.55          & 44.79                & \underline{54.28}          & 54.14          & 47.50                & 47.55          & \underline{51.68}    & 49.51          & 44.42       & 44.01 & \underline{50.52}    & 48.58          & 42.10       & 42.20 & 51.32          & \underline{52.44}    & 43.54       & 42.90 & \underline{53.78}    & 51.98       & 46.30       & 46.65       \\
                            & Gemma 4 31B       & \underline{58.02}    & 57.02          & 52.97          & 54.70                & \underline{61.27}          & 59.37          & 55.44                & 56.97          & \underline{57.63}    & 54.38          & 53.32       & 54.07 & \underline{56.04}    & 53.50          & 50.06       & 51.19 & \textbf{60.25} & 55.27          & 51.53       & 51.72 & \underline{59.07}    & 57.23       & 52.99       & 54.62       \\
                            & LLaMA 3.1 8B      & 10.71          & 17.61          & \underline{27.23}    & 21.78                & 7.70                 & 23.05          & \underline{36.60}          & 28.27          & 10.80          & 18.30          & \underline{32.17} & 27.74 & 10.08          & 14.78          & \underline{30.28} & 24.83 & 17.05          & 18.34          & \underline{30.78} & 24.06 & 8.33           & 24.26       & \underline{35.03} & 27.50       \\
                            & Qwen 3.6 27B          & \textbf{61.58} & 60.00          & 53.56          & 57.02                & \textbf{63.28}       & 62.98          & 57.35                & 62.00          & \textbf{60.12} & 59.09          & 54.97       & 57.39 & \textbf{57.18} & 56.28          & 51.13       & 53.78 & \underline{52.01}    & 49.94          & 45.48       & 49.06 & \textbf{61.80} & 59.99       & 56.91       & 58.19       \\
                            & Qwen 3.6 35B* & \underline{58.80}    & 56.54          & 51.65          & 54.59                & 58.94                & \underline{59.30}    & 53.55                & 55.46          & \underline{55.37}    & 55.08          & 50.67       & 51.69 & 50.73          & \underline{53.79}    & 50.93       & 50.53 & \underline{48.75}    & 47.93          & 41.41       & 42.32 & \underline{57.05}    & 56.40       & 53.60       & 54.88       \\
                            & Qwen 3.8 27B          & \underline{57.29}    & 54.74          & 46.82          & 50.90                & \underline{58.08}          & 57.42          & 51.09                & 54.30          & \underline{57.48}    & 51.82          & 47.86       & 50.15 & \underline{53.18}    & 51.54          & 44.01       & 47.47 & \underline{40.19}    & 37.30          & 34.42       & 37.65 & \underline{58.54}    & 54.56       & 50.24       & 53.31       \\
                                                         \cdashlinelr{2-26}
& \cite{wu-etal-2025-absa} &   \multicolumn{4}{c}{16.77} & \multicolumn{4}{c}{26.26} & \multicolumn{4}{c}{22.96} & \multicolumn{4}{c}{19.83} & \multicolumn{4}{c}{25.35} & \multicolumn{4}{c}{24.98} \\
                            \bottomrule
\end{tabular}
\end{adjustbox}
\label{tab:res}
\end{table*}

\subsection{Prompting Strategies}
To evaluate how instruction design and reasoning structures influence zero-shot cross-lingual TASD, we evaluate four prompting paradigms (Figure~\ref{fig:prompts_overview}):

\begin{enumerate}
    \item \textbf{Monolithic Extraction (Single-Pass Direct):} A rule-complete prompt specifying polarity/objectivity definitions, category constraints, and output formatting in a single pass without explicit prompt-level CoT instructions (Figure~\ref{fig:mono}), though models with native thinking modes still perform internal reasoning.
    
    \item \textbf{In-Context Chain-of-Thought (CoT):} A single-pass strategy instructing the model to emit explicit step-by-step reasoning prior to formatted tuples (Figure~\ref{fig:cot}), using concise definition blocks to reserve output budget for reasoning.
    
    \item \textbf{Sequential Pipeline (SQ):} A modular two-stage process (Steps 1--2 in Figure~\ref{fig:sequential}) that extracts candidate aspect terms before mapping them to categories and polarities; by construction, Step 1 bounds maximum achievable recall.
    
    \item \textbf{Sequential Refinement (SR):} A three-stage workflow (Steps 1--3 in Figure~\ref{fig:sequential}) adding a self-critique pass to verify predicted tuples against input substrings, valid categories, and polarities.
\end{enumerate}

\section{\uppercase{Results}}

Table~\ref{tab:res} presents the evaluation results. Across all non-English splits, top-performing zero-shot LLMs consistently outperform cross-lingual fine-tuned mT5 transfer, whereas in-language fine-tuned mT5 retains an advantage on English laptops ($43.54$ vs.\ $36.94$). In contrast, LLaMA~3.1~8B achieves substantially lower scores, likely due to its narrower multilingual pre-training (officially supporting eight languages) and lack of native reasoning capabilities. Notably, LLaMA~3.1~8B is the only model where multi-step prompting (SQ and SR) consistently improves over single-pass prompts, whereas monolithic prompting (MO) performs best across most larger architectures, closely followed by CoT.

Overall, the Qwen-3.6 family achieves the strongest performance across most configurations, outperforming Qwen-3.8-27B particularly on low-resource scripts such as Thai ($52.01$ vs.\ $40.19$ on restaurant). Performance remains stable across language families, demonstrating strong multilingual generalization. Across domains, restaurant yields the highest F1 scores, followed by hotel and laptop, reflecting increasing category granularity (13 categories in restaurant, 37 in hotel, and 114 in laptop).

These findings confirm that direct single-pass prompting is optimal for modern instruction-tuned LLMs in zero-shot multilingual ABSA, aligning with prior observations~\cite{wu2024evaluatingzeroshotmultilingualaspectbased}. Moreover, dense models consistently outperform MoE counterparts of comparable or larger total scale (e.g., Qwen-3.6-27B dense vs.\ Qwen-3.6-35B-A3B MoE). Because MoE models activate only 3B--4B parameters per token, dense architectures appear better equipped to handle simultaneous multilingual alignment, target extraction, and strict formatting constraints in zero-shot settings.

\subsection{Error Analysis}
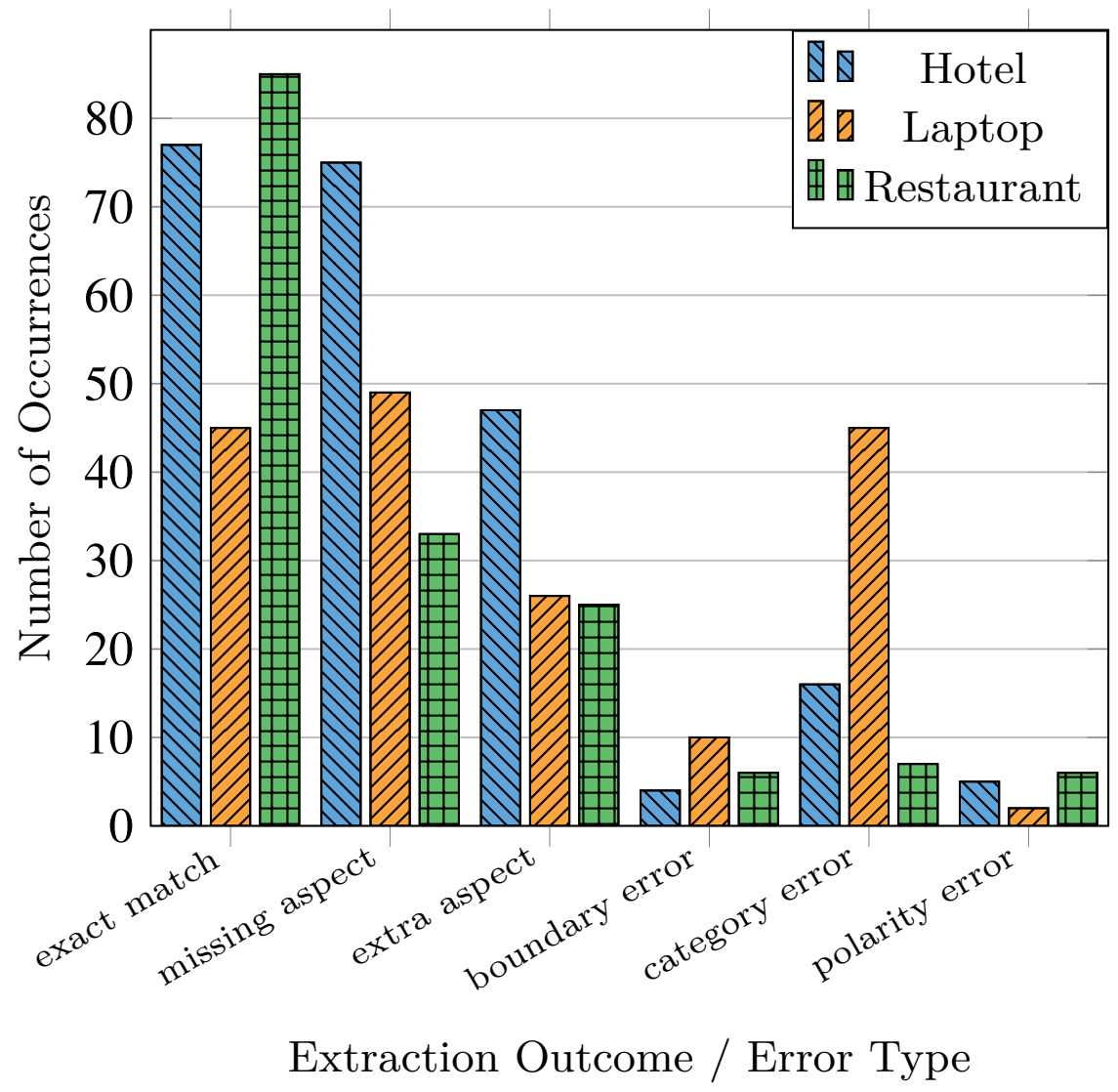
\begin{figure}[ht!]
        \centering
        \begin{adjustbox}{width=\linewidth}
            \begin{tikzpicture}
                \begin{axis}[
                    ybar,
                    bar width=8pt,
                    xtick={0,1,2,3,4,5},
                    ytick={0,10,20,30,40,50,60,70,80},
                    x tick label style={rotate=30, anchor=east, font=\scriptsize},
                    xticklabels={\scriptsize exact match,\scriptsize missing aspect,\scriptsize extra aspect,\scriptsize boundary error, \scriptsize category error, \scriptsize polarity error},
                    % symbolic x coords={exact match, missing aspect, extra aspect, boundary error, category error, polarity error},
                    ymin=0,
                    ymax=90,
                    xmin=-0.5,
                    xmax=5.5,
                    ymajorgrids=true,
                    ylabel={\footnotesize Number of Occurrences},
                    xlabel={\footnotesize Extraction Outcome / Error Type},
                    legend style={at={(1,1)}, anchor=north east, font=\footnotesize},
                    ]
                    \addplot[black,fill=lightblue,postaction={pattern=north west lines}] coordinates {(0,77) (1,75) (2,47) (3,4) (4,16) (5,5)};
                    \addplot[black,fill=lightorange,postaction={pattern=north east lines}] coordinates {(0,45) (1,49) (2,26) (3,10) (4,45) (5,2)};
                    \addplot[black,fill=lightgreen,postaction={pattern=grid}] coordinates {(0,85) (1,33) (2,25) (3,6) (4,7) (5,6)};
                    \legend{Hotel,Laptop,Restaurant}
                \end{axis}
            \end{tikzpicture}
        \end{adjustbox}
    \caption{Distribution of correct extractions and specific error types across evaluation domains for the Qwen 3.6 27B model using a monolithic prompt. Values represent the average occurrence count across 100 random test samples per language.}
    \label{fig:error_analysis}
\end{figure}

To gain deeper insight into the qualitative failure modes of our top-performing setup, we conducted a fine-grained error analysis on the output of the Qwen 3.6 27B model using the monolithic prompting strategy. For each evaluated domain, we randomly sampled 100 test instances per language and averaged occurrence counts across languages (note that exact matches plus missing aspects reflect per-draw gold aspect density across sampled subsets). The resulting error distribution across the three domains is illustrated in Figure~\ref{fig:error_analysis}.

We categorised model predictions into six distinct extraction outcomes:
\begin{enumerate}
    \item \textbf{Exact Match:} The predicted aspect boundary, category label, and sentiment polarity all align perfectly with the ground-truth tuple.
    \item \textbf{Missing Aspect:} A false negative scenario where a ground-truth aspect is completely omitted from the model's prediction.
    \item \textbf{Extra Aspect:} A false positive scenario where the model identifies an aspect target that does not exist in the gold annotations.
    \item \textbf{Boundary Error:} The predicted aspect span partially overlaps with the target aspect, but fails to capture its exact character boundaries.
    \item \textbf{Category Error:} The predicted category label is incorrect, evaluated strictly on aspect spans that achieved at least an exact or partial boundary match.
    \item \textbf{Polarity Error:} The predicted sentiment polarity is incorrect, evaluated strictly on aspect spans that achieved at least an exact or partial boundary match.
\end{enumerate}

As shown in Figure~\ref{fig:error_analysis}, complete omissions (missing aspects) constitute the single most frequent failure mode across all three evaluation domains. Conversely, fine-grained attribute classification errors are substantially less prevalent. Polarity errors are exceptionally rare across all domains, indicating that once a valid aspect candidate is identified, the model reliably assigns the correct sentiment direction. 

Category errors are similarly infrequent in the restaurant and hotel domains, but exhibit a marked increase within the laptop domain. This performance gap directly reflects the label space complexity of the respective schema: while the restaurant domain contains only 13 distinct aspect categories and the hotel domain contains 37, the laptop domain features a vastly larger taxonomy of 114 categories. The heightened ambiguity and fine-grained distinctions among laptop attribute types make accurate category assignment considerably more challenging. Finally, boundary errors remain relatively minor across all datasets, confirming that the model generally pinpoints span limits accurately when an aspect is successfully detected.

Furthermore, our output inspection revealed that certain architectures -- most notably LLaMA -- exhibited a pronounced sensitivity to output formatting and aspect targeting in zero-shot settings. In several instances, the model correctly identified categories and polarities but defaulted explicit targets to \textit{\quotes{null}}, or generated slightly malformed tuple string syntax that failed strict exact-match parsing rules. This formatting fragility directly accounts for localized drops in performance on specific subsets, underscoring the necessity of constrained decoding or flexible post-processing when evaluating some open LLMs.

\subsection{Step-Level Pipeline Dynamics and Refinement Analysis}

To analyze the internal mechanisms governing sequential decomposition, we examined step-level execution logs across LLaMA~3.1~8B, Qwen~3.6~27B, and Gemma~4~31B on a diagnostic subset of 100 English restaurant test sentences (121 gold triplets). Under deterministic greedy decoding, Step 1 extraction and Step 2 mapping are identical between SQ and SR, isolating the Step 3 refinement pass (Table~\ref{tab:pipeline_dynamics}).

The logs highlight two structural dynamics: First, initial extraction forms an unrecoverable recall ceiling by construction. Step 1 aspect recall scales with model capacity: Gemma~4~31B and Qwen~3.6~27B reach $74.38\%$ ($90/121$) and $72.73\%$ ($88/121$), whereas LLaMA~3.1~8B captures only $47.93\%$ ($58/121$). Because Steps 2 and 3 only process Step 1 candidates, omitted aspects cannot be recovered.

Second, refinement partially recovers multi-step losses on larger models but harms smaller ones. On Qwen~3.6~27B and Gemma~4~31B, Step 3 leaves $80.0\%$--$88.0\%$ of intermediate predictions unchanged, introduces $0.0\%$ degraded transitions (\textit{Harmed}), and corrects $2.0\%$--$3.0\%$ of Step 2 errors (\textit{Helped}) -- explaining why SR consistently improves over SQ in Table~\ref{tab:res} without reaching MO performance. Conversely, LLaMA~3.1~8B exhibits high volatility: Step 3 alters $57.0\%$ of outputs and degrades correct predictions ($9.0\%$ \textit{Harmed}) at more than twice the rate it fixes errors ($4.0\%$ \textit{Helped}).

These findings demonstrate that while multi-step decomposition offers necessary procedural scaffolding for smaller, non-reasoning architectures like LLaMA~3.1~8B, it degrades overall extraction quality on mid-scale models. Even when Step~3 refinement recovers a portion of these multi-step losses without harming valid predictions, the combined pipeline still underperforms single-pass monolithic extraction while incurring substantial computational overhead.

\begin{table}[ht!]
\centering
\caption{Step-level execution analysis of the sequential pipeline on a diagnostic subset of 100 English restaurant sentences (121 gold triplets). Step 1 reports aspect term recall; Step 3 measures refinement transitions relative to Step 2 predictions.}
\begin{adjustbox}{width=\linewidth}
\begin{tabular}{@{}lccccc@{}}
\toprule
\textbf{Model} & \textbf{Step 1 Aspect} & \multicolumn{4}{c}{\textbf{Step 3 Refinement Impact (\%)}} \\ \cmidrule(lr){3-6}
 & \textbf{Recall (\%)} & \textbf{Unchanged} & \textbf{Helped} & \textbf{Harmed} & \textbf{Modified (Incorrect)} \\ \midrule
LLaMA 3.1 8B & 47.93 & 43.0 & 4.0 & 9.0 & 44.0 \\
Qwen 3.6 27B & 72.73 & 88.0 & 3.0 & 0.0 & 9.0 \\
Gemma 4 31B & 74.38 & 80.0 & 2.0 & 0.0 & 18.0 \\ \bottomrule
\end{tabular}
\end{adjustbox}
\label{tab:pipeline_dynamics}
\end{table}

\subsection{Computational and Generation Overhead}
Real-world deployment also depends on generation overhead. We measure whitespace-delimited output words per sample (including native reasoning traces) as a hardware-agnostic proxy for generation volume across the 522 English restaurant test instances on Qwen~3.6~27B (Table~\ref{tab:generation_overhead}). Because native thinking mode is active, even the monolithic prompt generates $371.7$ words per sample, which explains its modest output difference compared to explicit CoT ($415.3$ words, $1.12\times$). Decomposing the task into multi-turn calls sharply inflates generation volume: sequential prompting requires $520.5$ words ($1.40\times$), while sequential refine reaches $885.1$ words ($2.38\times$) due to repeated reasoning across three calls without surpassing monolithic F1.

\begin{table}[ht!]
\centering
\caption{Generation volume and computational overhead across prompting strategies on the English restaurant split ($N=522$) using Qwen~3.6~27B. Output volume includes native reasoning traces and intermediate step outputs.}
\begin{adjustbox}{width=\linewidth}
\begin{tabular}{@{}lcc@{}}
\toprule
\textbf{Prompting Strategy} & \textbf{Avg. Output Words / Sample} & \textbf{Relative Generation Overhead} \\ \midrule
Monolithic & 371.7 & $1.00\times$ \\
Chain-of-Thought & 415.3 & $1.12\times$ \\
Sequential & 520.5 & $1.40\times$ \\
Sequential Refine & 885.1 & $2.38\times$ \\ \bottomrule
\end{tabular}
\end{adjustbox}
\label{tab:generation_overhead}
\end{table}

\section{\uppercase{Conclusion}}

In this paper, we present a systematic evaluation of zero-shot multilingual TASD across 12 languages and three domains using seven open-weight instruction-tuned LLMs. Our experiments demonstrate that large dense models paired with direct monolithic prompting achieve strong zero-shot performance, outperforming cross-lingual mT5 transfer across non-English splits without task-specific fine-tuning. Step-level and error analyses show that initial aspect recall bounds pipeline performance, multi-step decomposition degrades F1 relative to monolithic prompting, and iterative refinement only partially recovers multi-step losses on larger models while harming smaller models and incurring up to $2.38\times$ generation overhead. These findings suggest that direct, holistic prompt designs represent the most effective and computationally efficient paradigm for zero-shot aspect-based sentiment analysis in real-world deployments.

\section*{\uppercase{Acknowledgements}}
This work has been supported by the project R\&D of Technologies for Advanced Digitalization in the Pilsen Metropolitan Area (DigiTech) No. CZ.02.01.01/00/23\_021/0008436.
Computational resources were provided by the e-INFRA CZ project (ID:90254), supported by the Ministry of Education, Youth and Sports of the Czech Republic.

\bibliographystyle{apalike}
{\small
\bibliography{bibliography}}

\end{document}